\documentclass[journal]{IEEEtran}
\usepackage{cite}
\usepackage{amsmath,amssymb,amsfonts}
\usepackage{algorithmic}
\usepackage{graphicx}
\usepackage{textcomp}
\usepackage{colortbl}
\usepackage{xcolor}
\usepackage{multirow}
\usepackage{color}

\def\BibTeX{{\rm B\kern-.05em{\sc i\kern-.025em b}\kern-.08em
    T\kern-.1667em\lower.7ex\hbox{E}\kern-.125emX}}

\begin{document}
\title{WaveMan: mmWave-Based Room-Scale Human Interaction Perception for Humanoid Robots}

\author{Yuxuan Hu, Kuangji Zuo, Boyu Ma, Shihao Li, Zhaoyang Xia, \\ Feng Xu, \IEEEmembership{Senior Member, IEEE}, and Jianfei Yang, \IEEEmembership{Senior Member, IEEE}
\thanks{This work is supported by MOE Singapore Tier 1 Grant RG83/25, RS36/24 and a Start-up Grant from Nanyang Technological University. (Corresponding authors: Feng Xu, Jianfei Yang).}
\thanks{Yuxuan Hu, Zhaoyang Xia, and Feng Xu are with the Key Laboratory for Information Science of Electromagnetic Waves, Ministry of Education, School of Information Science and Technology, Fudan University, Shanghai 200433, China (e-mail: huyx23@m.fudan.edu.cn; xiazy@fudan.edu.cn; fengxu@fudan.edu.cn).}
\thanks{Yuxuan Hu is also with the School of Mechanical and Aerospace Engineering, Nanyang Technological University, Singapore 639798, as a visiting Ph.D. student. }
\thanks{Kuangji Zuo, Boyu Ma, Shihao Li, and Jianfei Yang are with the School of Mechanical and Aerospace Engineering, Nanyang Technological University, Singapore 639798  (e-mail: kuangji001@e.ntu.edu.sg; boyu.ma@ntu.edu.sg; n2409375d@e.ntu.edu.sg; jianfei.yang@ntu.edu.sg).}
}

\maketitle

\begin{abstract}
Reliable humanoid-robot interaction (HRI) in household environments is constrained by two fundamental requirements, namely robustness to unconstrained user positions and preservation of user privacy. Millimeter-wave (mmWave) sensing inherently supports privacy-preserving interaction, making it a promising modality for room-scale HRI. However, existing mmWave-based interaction-sensing systems exhibit poor spatial generalization at unseen distances or viewpoints. To address this challenge, we introduce WaveMan, a spatially adaptive room-scale perception system that restores reliable human interaction sensing across arbitrary user positions. WaveMan integrates viewpoint alignment and spectrogram enhancement for spatial consistency, with dual-channel attention for robust feature extraction. Experiments across five participants show that, under fixed-position evaluation, WaveMan achieves the same cross-position accuracy as the baseline with five times fewer training positions. In random free-position testing, accuracy increases from 33.00\% to 94.33\%, enabled by the proposed method. These results demonstrate the feasibility of reliable, privacy-preserving interaction for household humanoid robots across unconstrained user positions.
\end{abstract}




\section{Introduction}
\IEEEPARstart{H}{umanoid} robots have achieved substantial progress in perception, manipulation, and control in recent years~\cite{1024.1}. However, in human-centered interaction scenarios, reliably understanding human intent while preserving user privacy remains a key bottleneck hindering their deployment in everyday environments. In real indoor settings, users may initiate interaction from arbitrary locations within a room, invalidating traditional assumptions of front-facing, short-range, or user-aligned interaction. Consequently, intent understanding must remain reliable under significant variations in distance, viewpoint, and spatial relationships~\cite{1021.0}.

\begin{figure}[!t]\centering
	\includegraphics[width=8.5cm]{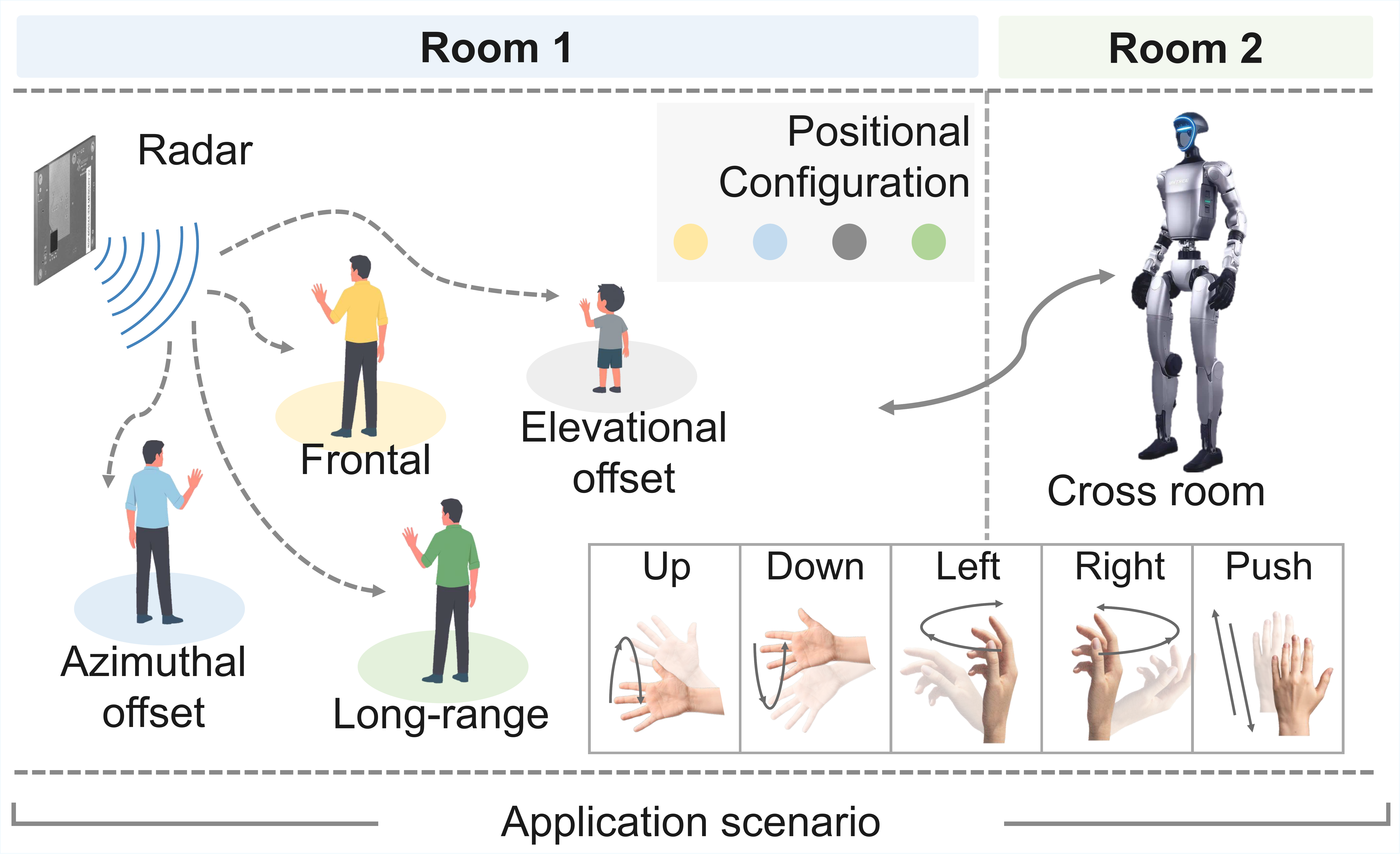}
	\caption{\textbf{Spatially adaptive room-scale interaction scenario.} WaveMan aligns observations from different user spatial positions into a unified perception space to mitigate spatial inconsistencies. 
    }\label{fig_1}
\end{figure}

Gestures constitute an intuitive human--robot interaction modality that conveys intent without speech or wearable devices, making them suitable for natural interaction in unconstrained spatial settings~\cite{1020.6}. However, most existing gesture recognition systems are designed for short-range scenarios with fixed viewpoints, limiting their applicability in real environments~\cite{1024.3}. In unconstrained human--robot interaction scenarios, the system cannot assume control over user pose or sensing viewpoint, and thus must rely on passively acquired observations from a fixed sensor. Accordingly, we consider an ambient sensing setting in which an environment-installed mmWave radar (e.g., a smart-home device) provides privacy-preserving observations to the robot for interaction. Under this setting, the relative distance, azimuth, and elevation between the user and the sensing node can vary substantially, imposing stricter requirements on the stability of room-scale gesture recognition across positions and viewpoints, as illustrated in Fig.~\ref{fig_1}.

Millimeter-wave (mmWave) radar has recently emerged as a promising sensing modality for room-scale gesture recognition~\cite{4}. Its non-intrusive sensing capability, robustness to lighting conditions and occlusion, and inherent privacy-preserving properties make it particularly suitable for long-term deployment in domestic environments to support humanoid interaction, offering clear advantages over vision-based~\cite{1024.0}, audio-based~\cite{1019.1}, and contact-based~\cite{1024.2} systems. Prior studies have demonstrated that mmWave radar can capture fine-grained hand motion dynamics with high temporal resolution~\cite{10}. 
However, most existing mmWave-based gesture recognition methods remain limited to short-range or frontal configurations~\cite{122}, resulting in poor robustness under room-scale interaction conditions. Although recent efforts have explored long-range mmWave gesture recognition~\cite{15}, many of these approaches rely on raw ADC data, incurring high data throughput and bandwidth requirements that hinder practical deployment on resource-constrained in-home embedded sensing nodes or edge hubs.


Point-cloud-based mmWave methods leverage chip-level preprocessing to extract more compact spatial representations, partially alleviating system-level data throughput and bandwidth constraints~\cite{171,20}. Existing approaches typically follow two paradigms: (i) end-to-end learning directly on raw or lightly processed point cloud sequences~\cite{171}, and (ii) transforming point cloud sequences into denser representations, such as time--frequency spectrograms, for deep learning~\cite{20}. Despite notable progress~\cite{23}, room-scale gesture recognition remains challenging. Substantial variations in user spatial position cause the same gesture to exhibit highly inconsistent observations across different distances and viewpoints, leading to pronounced feature distribution shifts and degraded generalization performance under unseen positions or viewpoints~\cite{310,34}. Moreover, gestures performed at long distances or off-axis directions typically produce weaker and sparser echoes with lower signal-to-noise ratios, further impairing the reliability of spectrogram-based representations~\cite{26,301}. Together, these factors induce geometry-, spectrum-, and direction-domain distribution shifts driven by spatial position changes, making existing methods ineffective under the random spatial configurations in domestic environments where the radar viewpoint is fixed. As a result, room-scale mmWave-based gesture recognition remains a largely unresolved challenge.

To address these challenges, this paper proposes a spatially adaptive gesture recognition framework tailored for room-scale human--robot interaction. The main contributions of this work are summarized as follows:
\begin{enumerate}
\item \textbf{A spatially adaptive room-scale perception framework for humanoid robot interaction.}  
We develop a room-scale gesture interaction system for humanoid robots that integrates mmWave sensing with robot behavior execution, enabling stable interaction performance across varying user positions.

\item \textbf{Geometry-, spectrum-, and direction-aware representation learning.}  
We propose a spatially adaptive processing pipeline to address key spatial variations induced by user position changes in room-scale interaction scenarios. The pipeline consists of: (i) geometric alignment to compensate for position and viewpoint variations; (ii) unsupervised spectrogram enhancement to mitigate long-range sparsity and low signal-to-noise ratios; and (iii) an attention-based recognition network that fuses multi-channel information to improve robustness across positions.

\item \textbf{Room-scale deployment and real-world humanoid robot evaluation.}  
The proposed framework is deployed on a humanoid robot platform and evaluated under fixed-position, unseen-position, and random-position settings. Experimental results demonstrate reliable whole-room intent understanding, stable robot behavior execution, and significant performance improvements over existing methods.
\end{enumerate}

\begin{figure*}[!t]
    \centering
    \includegraphics[width=17.5cm]{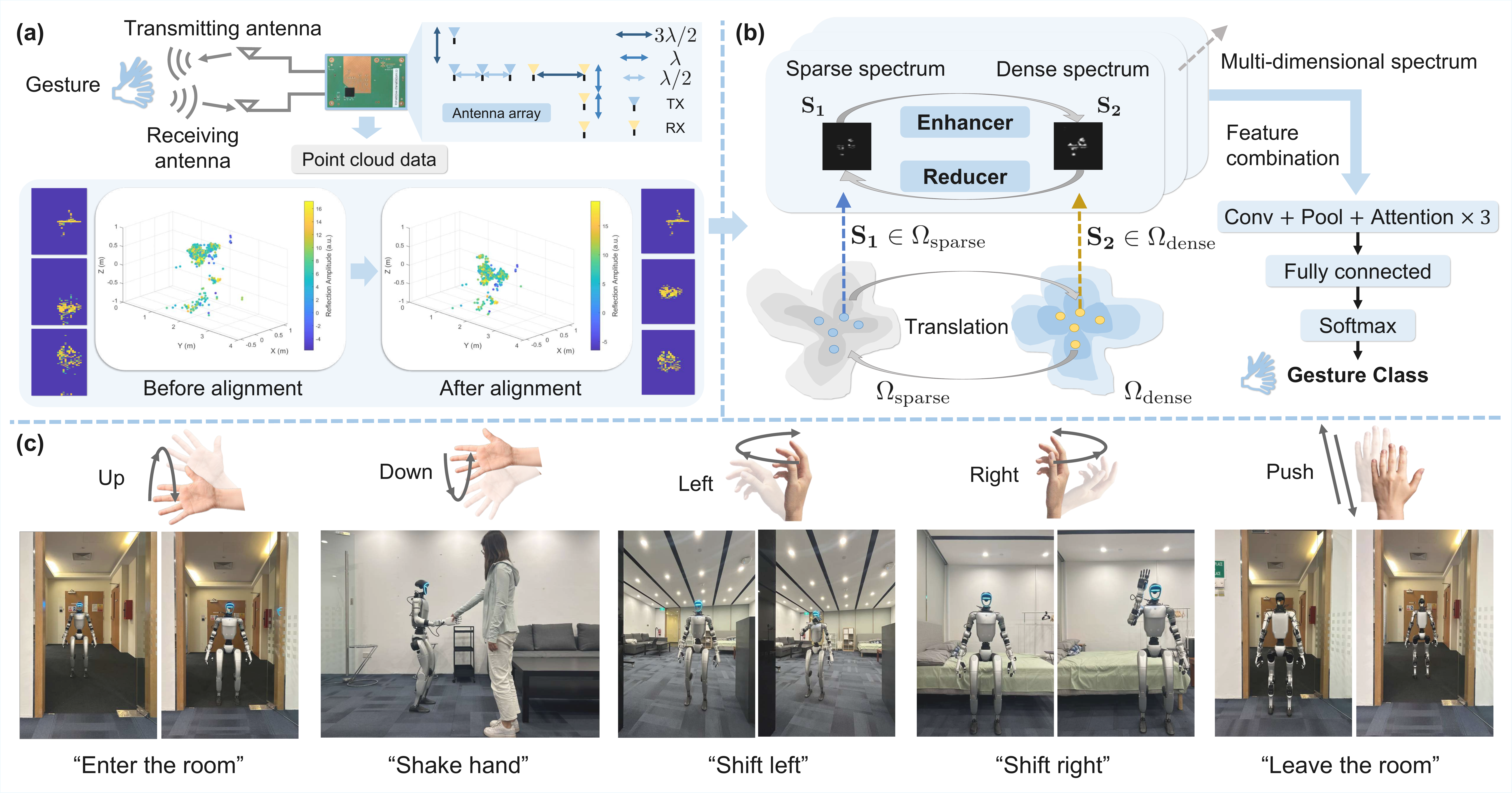}
    \caption{\textbf{Overview of the proposed spatially adaptive interaction framework.} 
(a) Radar point-cloud data captured under diverse positional configurations are spatially aligned and transformed into spectrogram representations.
(b) Sparse spectrograms are enhanced and fused with dense spectra to obtain robust gesture representations.
(c) The recognized gestures are mapped to corresponding humanoid robot behaviors, enabling reliable human--robot interaction across unconstrained user positions.
}
    \label{fig_2}
\end{figure*}

An overview of the entire framework is presented in Fig.~\ref{fig_2}. 
The remainder of this paper is organized as follows: 
Section~II reviews related work; 
Section~III introduces the mmWave radar signal model and spectrogram derivation; 
Section~IV presents the proposed spatially adaptive perception framework; 
Section~V describes experimental setups and results; 
and Section~VI concludes the paper.

\section{Related Work}

\subsection{Gesture-based Human--Robot Interaction}

Gesture-based interaction has become an important modality for natural and contact-free human-robot communication~\cite{1023.0}. Millimeter-wave (mmWave) radar is particularly appealing for HRI due to its robustness to lighting variations, capability of capturing fine-grained motion, and strong privacy preservation~\cite{1023.2}. Prior systems based on FMCW and impulse radars have demonstrated reliable gesture-driven control of robots in structured environments~\cite{10}. However, most radar-based gesture recognition frameworks are developed for short-range and fixed-view setups, where the relative user-radar geometry remains stable~\cite{122}. These constraints limit their applicability to room-scale scenarios, in which users can move freely and perform gestures from diverse spatial positions and orientations. Such settings introduce two fundamental challenges: spatial variation causing position-dependent feature inconsistency, and signal degradation resulting in sparse or incomplete spectrograms under long-range or off-axis conditions. These challenges motivate the need for radar-based HRI systems that can operate robustly across wide spatial configurations, as elaborated in the following subsections.

\subsection{Spatial Robustness in Gesture Recognition}

The spatial relationship between the user and radar varies substantially in room-scale environments, leading to geometric distortions and inconsistent feature representations that impair recognition under unseen viewpoints. Several learning-based strategies have been proposed to mitigate this issue. Zhang et al.~\cite{32} employed a convolutional autoencoder to extract position-invariant gesture features, while Li et al.~\cite{310} utilized signal-aware data augmentation to improve generalization across unseen spatial configurations. Xia et al.~\cite{20} introduced a viewpoint alignment scheme to normalize multi-view gesture data, and Li et al.~\cite{34} applied domain adaptation to reduce distribution gaps caused by varying user positions. In addition, a ShuffleNet-based system has been shown to achieve high accuracy on 43 alphanumeric and symbolic gestures using a single radar unit~\cite{341}. Although these approaches improve model-level robustness, most operate at the feature level and do not explicitly standardize the geometric structure of the radar input. Consequently, variations in user-radar distance, azimuth, and elevation can still lead to inconsistent spatial patterns, limiting robustness in unseen room-scale conditions.

\subsection{Sparse Gesture Representations}

Long-range or off-axis gestures often produce sparse and low-SNR spectrograms due to weak reflections and incomplete signal returns, posing another major challenge for room-scale recognition. Efforts to address this problem include both network-level and data-level enhancement strategies. Dong et al.~\cite{26} proposed spatiotemporal deformable convolution and context-aware modules to improve robustness against degraded inputs, while Towakel et al.~\cite{27} designed a multimodal attention mechanism to enhance temporal fusion. Data-centered methods such as spectrogram inpainting, physically interpretable augmentation~\cite{30}, and deep-learning-based reconstruction~\cite{301} have also shown promise in restoring missing or attenuated signal components. However, these methods often target specific degradation patterns or operate under fixed spatial configurations, making them less effective when degradation is tightly coupled with spatial variation. This limitation highlights the need for adaptive enhancement mechanisms that can account for user-radar geometry and provide consistent representations across room-scale environments.

\section{Radar Signal Model and Spectrogram Derivation}

This section summarizes the signal processing principles of the MIMO FMCW radar employed in this work, which serves as the primary sensing modality for humanoid robot interaction in indoor environments.

\subsection{FMCW Signal Model}

The transmitted waveform of a linear frequency-modulated continuous-wave (LFMCW) radar during a single chirp is  

\begin{equation}
S_{\mathrm{tx}}(t) = A_{\mathrm{T}} \exp\!\left[-j 2\pi\!\left(f_0 t + \frac{B}{2T_{\mathrm{s}}} t^2 \right)\right], \quad 0 \le t \le T_{\mathrm{s}},
\end{equation}
where \(A_{\mathrm{T}}\) is the amplitude, \(f_0\) is the carrier frequency, \(B\) is the bandwidth, and \(T_{\mathrm{s}}\) is the chirp duration.  
A target at range \(R\) and radial velocity \(v\) returns a delayed and Doppler-shifted copy of the transmitted chirp. Mixing the received and transmitted signals followed by low-pass filtering produces the intermediate-frequency (IF) signal
\begin{equation}
S_{\mathrm{if}}(t) = A_{\mathrm{T}} A_{\mathrm{R}} \exp\!\left[j (\phi_{\mathrm{tx}}(t) - \phi_{\mathrm{rx}}(t)) \right],
\end{equation}
whose instantaneous phase encodes both range and velocity.  

A fast-time FFT across samples within each chirp yields the range spectrum, while a slow-time FFT across consecutive chirps extracts Doppler information. Combining the two produces a two-dimensional range-Doppler (RD) map. The range and velocity resolutions follow standard FMCW relationships~\cite{35}, i.e.,  
\(\Delta r = c/(2B)\) and \(\Delta v = \lambda /(2 N_{\mathrm{chirp}} T_{\mathrm{s}})\).

\subsection{Angle Estimation and 5D Point-Cloud Representation}

Spatial directionality is obtained via digital beamforming (DBF) on the MIMO virtual array. One-dimensional DBF estimates azimuth, while two-dimensional DBF jointly estimates azimuth and elevation, forming a three-dimensional angle–elevation (AE) spectrum.

Each radar frame produces a set of scattering points represented as
\begin{equation}
P = \left\{ \big(\tilde{A}_i, r_i, v_i, h_i, e_i\big) \mid i=1,\ldots,N_{\mathrm{P}} \right\},
\end{equation}
where $\tilde{A}_i$ denotes the complex amplitude, and $r_i$, $v_i$, $h_i$, and $e_i$ are the range, radial velocity, azimuth, and elevation of the $i$-th reflection, respectively. The angular components are converted to Cartesian coordinates $(x_i,y_i,z_i)$, yielding a 3D radar point cloud that serves as the geometric input for subsequent spatial alignment.

\subsection{Multi-Domain Time-Spectrogram Construction}

To characterize gesture dynamics over time, multiple time-spectrograms are constructed across different physical domains, including range–time (RT), Doppler–time (DT), azimuth–time (HT), elevation–time (ET), and position–time (XT, YT, ZT) representations derived from Cartesian coordinates.

These multi-domain spectrograms provide complementary descriptions of gesture evolution in range, velocity, direction, and spatial displacement, and are used as structured inputs to the proposed spatially adaptive perception modules.

\section{Spatially Adaptive Perception Framework}

\subsection{System Overview and Robot Integration}

This section introduces the proposed spatially adaptive perception framework for room-scale human--robot interaction in humanoid robot systems, which explicitly accounts for unconstrained user positions and viewpoints (Fig.~\ref{fig_2}). The framework processes streaming mmWave radar measurements through a unified pipeline for room-scale humanoid interaction, including spatially adaptive point cloud alignment, spectrogram-based feature construction and enhancement, and viewpoint-adaptive recognition with dual-channel attention to infer interaction gestures. 

The perception framework runs on an external workstation, while interaction control is executed on the humanoid robot’s onboard controller, with inferred gesture labels and confidence transmitted via UDP to form a closed perception--action loop. During operation, the system runs online in a streaming manner, producing interaction predictions over temporal windows and enabling consistent robot responses across room-scale distances and viewpoints. All experimental evaluations in Section~V follow this setup.

\begin{figure}[!t]
    \centering
    \includegraphics[width=8.5cm]{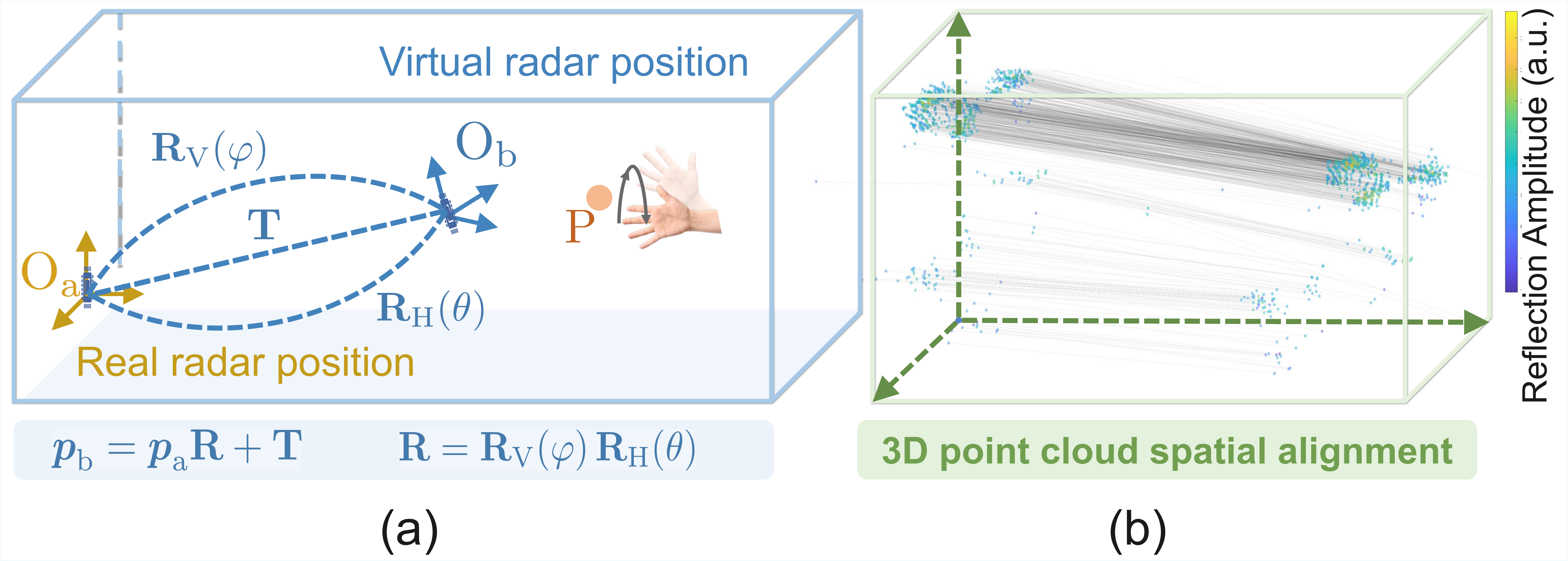}
    \caption{\textbf{Spatially adaptive point cloud alignment.}
    (a) shows the geometric relationship between the real radar position $\mathrm{O_a}$, the virtual canonical radar position $\mathrm{O_b}$, and the observed point $\mathrm{P}$, where azimuth and elevation offsets are compensated through sequential rotations and translation.
    (b) presents an example point cloud before and after alignment, illustrating reduced angular distortion and a more compact spatial distribution for subsequent spectrogram generation.}
    \label{fig_3}
\end{figure}

\begin{figure*}[!t]
    \centering
    \includegraphics[width=17.5cm]{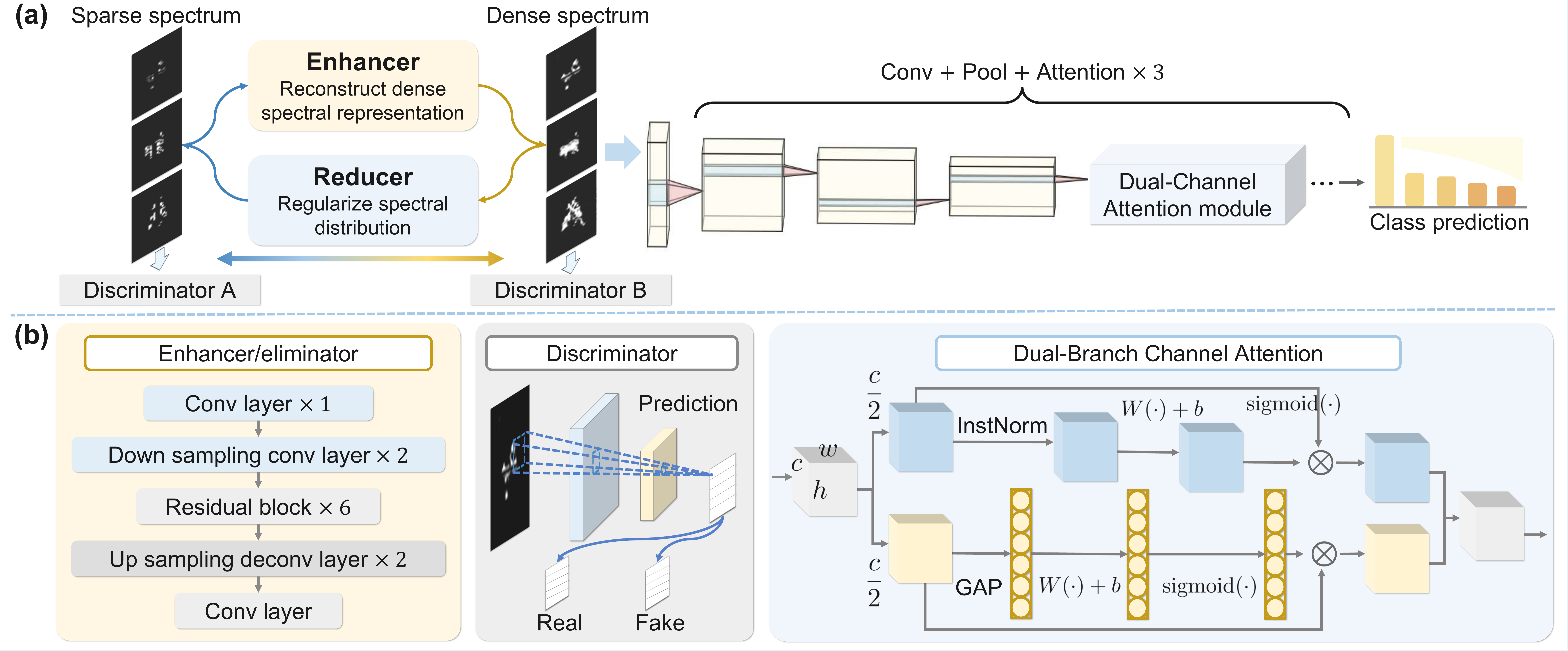}
    \caption{\textbf{Architecture of the proposed spectrogram enhancement and recognition network.}
(a) Unpaired spectrogram enhancement pipeline, where sparse spectral inputs are translated into dense representations using an Enhancer–Reducer pair supervised by two discriminators.
(b) Internal network components, including the Enhancer/Reducer architectures, PatchGAN-based discriminators, and a compact CNN equipped with the proposed Dual-Branch Channel Attention (DBCA) module for generalization-oriented gesture recognition.}
    \label{fig_5}
\end{figure*}

\subsection{Spatially Adaptive Point Cloud Alignment}

Non-frontal radar perspectives introduce significant geometric distortions in raw point clouds, including spatial compression, angular skew, and amplitude attenuation due to oblique reflections. Moreover, the nonlinear mapping from spherical measurements to Cartesian coordinates further amplifies viewpoint-dependent inconsistencies. To unify the appearance of gestures observed at different distances and orientations, we propose a spatially adaptive point cloud alignment algorithm that reprojects each aggregated gesture instance into a canonical front-facing configuration, as shown in Fig.~\ref{fig_3}.

\subsubsection{Motivation and Geometric Deformation}

When a user performs gestures at nonzero elevation or azimuth relative to the antenna boresight, the radar observes a geometrically distorted point cloud. This distortion arises from oblique-angle surface projection and nonlinear spherical-to-Cartesian mapping in the radar measurement model, causing identical gestures to exhibit inconsistent spatial distribution, density, and energy across different user--radar configurations. The goal of spatial alignment is therefore to suppress such viewpoint-induced variations by transforming each point cloud into a canonical coordinate system corresponding to a frontal user orientation.

\subsubsection{Canonical Reprojection Model}

Let the superscript \({}^{(j)}\) denote the \(j\)-th aggregated point cloud of a gesture instance. Subscripts~\(\mathrm{a}\) and \(\mathrm{b}\) denote the original and aligned coordinate systems, respectively. The transformation from \(\mathrm{a}\) to \(\mathrm{b}\) jointly compensates elevation and azimuth offsets, performs distance normalization, and corrects amplitude attenuation. The reprojection model is defined as
\begin{equation}
\label{eq:align_with_proj}
\begin{aligned}
r_{\mathrm{b}}^{(j)} &= r_{\mathrm{a}}^{(j)} \cos\!\big(\varphi^{(j)}\big)\cos\!\big(\theta^{(j)}\big), \\[3pt]
\varphi_{\mathrm{b}}^{(j)} &= \arctan\!\big(y_0 \tan(\varphi_{\mathrm{a}}^{(j)} - \varphi^{(j)})\big), \\[3pt]
\theta_{\mathrm{b}}^{(j)} &= \arctan\!\big(y_0 \tan(\theta_{\mathrm{a}}^{(j)} + \theta^{(j)})\big), \\[3pt]
A_{\mathrm{b}}^{(j)} &= 
\dfrac{A_{\mathrm{a}}^{(j)}}%
{\cos\!\big(\varphi^{(j)}\big)\cos\!\big(\theta^{(j)}\big)}.
\end{aligned}
\end{equation}

Here, \(\varphi^{(j)}\) and \(\theta^{(j)}\) are the estimated elevation and azimuth deviations, and \(y_0\) is a reference distance constant used to stabilize the virtual reprojection. This formulation approximates a mapping to a virtual frontal viewpoint through empirical calibration.
The reprojection standardizes the geometry and amplitude distribution of the point cloud, ensuring a unified spatial structure for gestures observed from different viewpoints.

\subsubsection{Real-Time Parameter Estimation}

To enable adaptive alignment in real time, the elevation offset \(\varphi^{(j)}\), azimuth offset \(\theta^{(j)}\), and mean vertical distance \(\bar y^{(j)}\) are estimated directly from radar spectrograms. Specifically, range-time (RT), elevation-time (ZT), and azimuth-time (XT) spectrograms are used to extract radial, vertical, and horizontal motion cues, respectively. Let \(\mathrm{O_aP}^{(j)}\) denote the radial distance between the radar origin \(\mathrm{O_a}\) and the representative scattering point~\(P\), and let \(\mathrm{O_aP_v}^{(j)}\) and \(\mathrm{O_aP_h}^{(j)}\) denote its vertical and horizontal displacements. The geometric relationships yield
\begin{equation}
\label{eq:angle_est}
\begin{aligned}
\varphi^{(j)} &= 
\arcsin\!\left(
\frac{\mathrm{O_aP_v}^{(j)}}{\mathrm{O_aP}^{(j)}}
\right), \\[3pt]
\theta^{(j)} &= 
\arcsin\!\left(
\frac{\mathrm{O_aP_h}^{(j)}}{\mathrm{O_aP}^{(j)}}
\right).
\end{aligned}
\end{equation}

The radial distance is estimated from the RT spectrogram using an amplitude-weighted centroid:
\begin{equation}
\label{eq:rt_centroid}
\mathrm{O_aP}^{(j)} = 
\frac{1}{N_{\text{thre-r}}}
\sum_{t=1}^{N_F} \sum_{i=1}^{N_\text{adc}/2} 
i \, r_{\text{res}} \, f(R_t(i)),
\end{equation}
where \(r_{\text{res}}\) is the range resolution and

\begin{equation}
f(x) = 
\begin{cases}
1, & x > A_{\text{thre-r}}, \\
0, & \text{otherwise}.
\end{cases}
\end{equation}

Vertical and horizontal displacements are obtained analogously from the ZT and XT spectrogram centroids.  
The estimated mean vertical distance
\begin{equation}
\bar y^{(j)} = 
\frac{1}{N}
\sum_{n=1}^{N}
r_a^{(j)}(n)\cos\!\big(\varphi_a^{(j)}(n)\big)
\cos\!\big(\theta_a^{(j)}(n)\big)
\end{equation}
is used to determine whether projection normalization is activated. When \(\bar y^{(j)} > \tau_y\), the transformation in \eqref{eq:align_with_proj} is applied using the stabilized reference distance \(y_0\).

\subsubsection{Noise Suppression and Center Normalization}

After spatial alignment, density-based noise suppression and center normalization are applied to refine the spectrogram representation~\cite{DBSCAN}. The denoising step removes low-density reflections unrelated to the main gesture region, while center normalization shifts the dominant cluster to the geometric center, improving spatial consistency across frames. These refinements yield a compact and well-centered spectrogram representation, facilitating more consistent gesture feature extraction.

\begin{figure*}[!t]
    \centering
    \includegraphics[width=17.5cm]{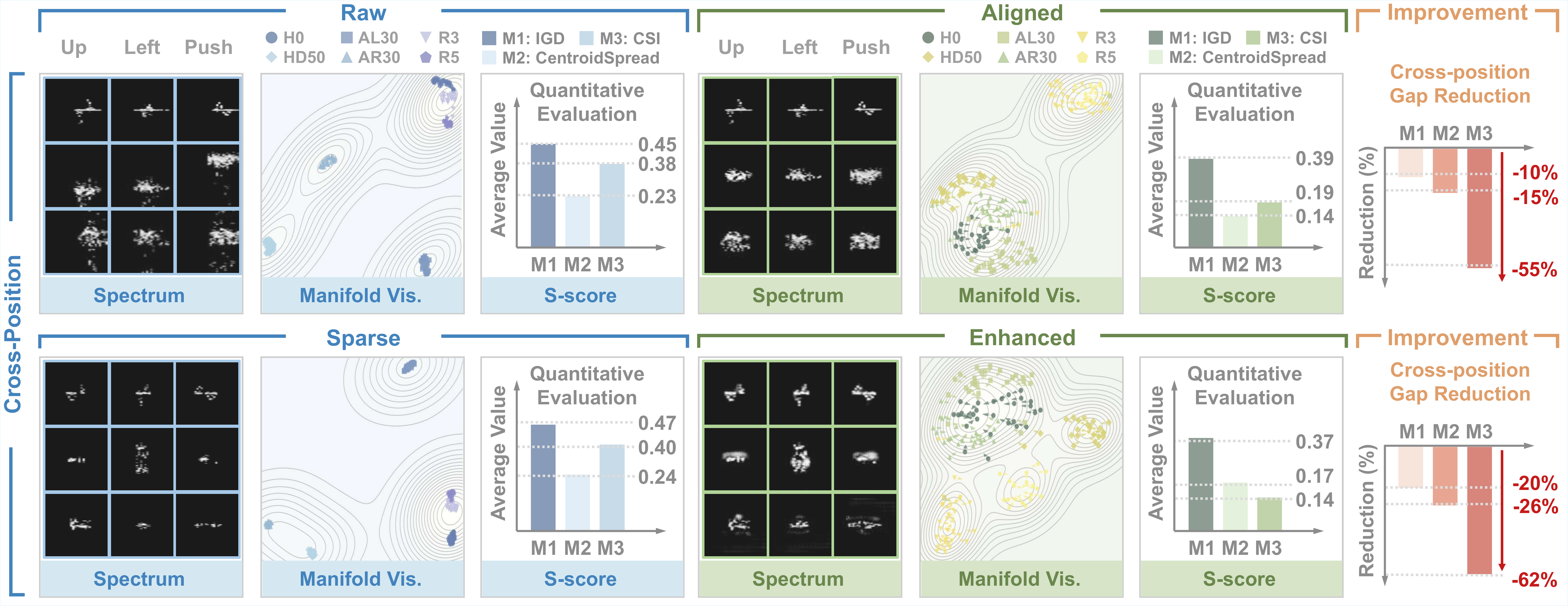}
\caption{Sampled examples of spectrograms and quantitative metrics before/after alignment (top) and enhancement (bottom).}
    \label{fig_4}
\end{figure*}

\subsection{Spectrogram Construction and Enhancement}

Following spatial alignment, radar frames captured during humanoid robot interaction are converted into the multi-domain temporal--spectral representations introduced in Section~III-C.
These spectrograms are subsequently enhanced using an unpaired spectral translation module to mitigate long-range and oblique-view degradation and to recover more complete spectral patterns for recognition.

The overall enhancement--recognition framework is illustrated in Fig.~\ref{fig_5}, where sparse spectrograms are first enhanced and then fed into a recognition network. Long-range and off-axis humanoid interaction scenarios often yield sparse or fragmented spectrograms due to signal attenuation and weak backscattering. To reconstruct dense and continuous spectral patterns without requiring labels, we adopt an unpaired spectral domain translation framework inspired by CycleGAN~\cite{41}.

As shown in Fig.~\ref{fig_5}(a), the enhancement module uses an Enhancer and a Reducer to reconstruct dense spectral patterns from sparse observations, supervised by two discriminators that operate on the sparse and dense domains. Let the sparse and dense spectrogram domains be denoted by 
\(\mathcal{S}_1\) and \(\mathcal{S}_2\), with samples 
\(\mathbf{S}_1 \sim p_{\mathcal{S}_1}\) and 
\(\mathbf{S}_2 \sim p_{\mathcal{S}_2}\).  
Two generators—the forward translator 
\(E: \mathcal{S}_1 \rightarrow \mathcal{S}_2\) 
and backward translator 
\(R: \mathcal{S}_2 \rightarrow \mathcal{S}_1\)-learn bidirectional mappings regulated by adversarial and cycle-consistency losses:
\begin{equation}
\label{eq:overall_loss_new}
\begin{aligned}
L(E, R, A, B) &=
L_{\text{GAN}}(E, B, \mathcal{S}_1, \mathcal{S}_2) 
+ L_{\text{GAN}}(R, A, \mathcal{S}_2, \mathcal{S}_1) \\
&\quad + \lambda\, L_{\text{cyc}}(E, R),
\end{aligned}
\end{equation}

\begin{equation}
\label{eq:cycle_loss_new}
\begin{aligned}
L_{\text{cyc}}(E, R) =
&\mathbb{E}_{\mathbf{S}_1}
\!\left[\|R(E(\mathbf{S}_1)) - \mathbf{S}_1\|_1\right]  \\
&\quad + 
\mathbb{E}_{\mathbf{S}_2}
\!\left[\|E(R(\mathbf{S}_2)) - \mathbf{S}_2\|_1\right].
\end{aligned}
\end{equation}

Both generators adopt a residual encoder-decoder architecture~\cite{43}, while each discriminator uses a \(34\times34\) PatchGAN to enforce local spectral realism. Training stability is improved using a replay buffer, which stores previously generated samples for discriminator updates. The resulting enhanced spectrograms exhibit improved continuity, stronger structural cues, and reduced sparsity, providing more reliable inputs for humanoid robot interaction.

\subsection{Generalization-Oriented Recognition Network}

The enhanced spectrograms are fed into an attention-based recognition network that extracts discriminative spectral features robust to spatial variations. The network employs a compact CNN backbone with a Dual-Branch Channel Attention (DBCA) mechanism to reweight spectral features that remain stable across user positions in room-scale humanoid interaction.

\subsubsection{Dual-Branch Channel Attention Architecture}

Standard CNNs treat feature channels uniformly, which can attenuate informative spectral cues under spatially varying observation conditions. To address this issue, we introduce a Dual-Branch Channel Attention (DBCA) module inspired by squeeze-and-excitation~\cite{SENET}, as illustrated in Fig.~\ref{fig_5}(b). 

DBCA constructs two complementary descriptor branches to capture local texture patterns and global motion signatures. The descriptors are fused and transformed to generate channel-wise attention weights, which adaptively rescale intermediate feature maps to enhance informative responses and suppress irrelevant activations.

\subsubsection{Gesture Classification Module}

The backbone comprises three convolution–batch-normalization–ReLU–max-pooling blocks, followed by two fully connected layers and a softmax classifier. Dropout is applied after each block. The network is trained using Adam with an initial learning rate of 0.001, a batch size of 64, and 70 epochs.

The complete enhancement–recognition pipeline, summarized in Fig.~\ref{fig_5}, robustly handles long-range degradation and spatial variation in room-scale interaction scenarios. The pipeline occupies 124.6~MB and achieves an average inference time of 5.45 ms per sample, demonstrating suitability for real-time humanoid–robot interaction.

\begin{table*}[t]
\centering
\caption{
Cross-position recognition performance under six training configurations.
All values are accuracies (\%). 
$|\mathcal{P}_{\mathrm{train}}|$ denotes the number of training positions (1--6).
The corresponding training sets are:
$|\mathcal{P}_{\mathrm{train}}|=1$: $\{\mathrm{P1}\}$;
$|\mathcal{P}_{\mathrm{train}}|=2$: $\{\mathrm{P1},\mathrm{P3}\}$;
$|\mathcal{P}_{\mathrm{train}}|=3$: $\{\mathrm{P1},\mathrm{P3},\mathrm{P5}\}$;
$|\mathcal{P}_{\mathrm{train}}|=4$: $\{\mathrm{P1},\mathrm{P3},\mathrm{P5},\mathrm{P6}\}$;
$|\mathcal{P}_{\mathrm{train}}|=5$: $\{\mathrm{P1},\mathrm{P3},\mathrm{P4},\mathrm{P5},\mathrm{P6}\}$;
$|\mathcal{P}_{\mathrm{train}}|=6$: $\{\mathrm{P1},\dots,\mathrm{P6}\}$.
“B/O” denotes Baseline vs.\ Ours.
}
\label{tab:CrossPosition1}
\renewcommand{\arraystretch}{1.15}
\setlength{\tabcolsep}{4pt}

\begin{tabular}{c |
cc cc cc cc cc cc |
cc |
c >{\columncolor{gray!20}}c |
c >{\columncolor{gray!20}}c}
\hline
\multirow{2}{*}{$|\mathcal{P}_{train}|$}
& \multicolumn{12}{c|}{Per-position Accuracy (B/O)}
& \multicolumn{2}{c|}{Seen-Mean}
& \multicolumn{2}{>{\columncolor{gray!20}}c|}{Unseen-Mean $\uparrow$}
& \multicolumn{2}{>{\columncolor{gray!20}}c}{Gap $\downarrow$} \\
\cline{2-19}
&
P1$_{\mathrm{B}}$ & P1$_{\mathrm{O}}$
& P2$_{\mathrm{B}}$ & P2$_{\mathrm{O}}$
& P3$_{\mathrm{B}}$ & P3$_{\mathrm{O}}$
& P4$_{\mathrm{B}}$ & P4$_{\mathrm{O}}$
& P5$_{\mathrm{B}}$ & P5$_{\mathrm{O}}$
& P6$_{\mathrm{B}}$ & P6$_{\mathrm{O}}$
& B & O
& B & O
& B & O \\
\hline

1
& 97.66 & 98.03
& 58.94 & 96.02
& 50.74 & 68.90
& 68.54 & 78.23
& 68.93 & 79.80
& 55.69 & 78.80
& 97.66 & 98.03
& 60.57 & \cellcolor{gray!20}{\textbf{80.35}}
& 37.09 & \cellcolor{gray!20}{\textbf{17.68}} \\

2
& 97.66 & 98.28
& 67.53 & 93.83
& 98.45 & 99.37
& 58.52 & 91.50
& 80.13 & 78.01
& 65.72 & 85.44
& 98.05 & 98.82
& 67.97 & \cellcolor{gray!20}{\textbf{87.20}}
& 30.08 & \cellcolor{gray!20}{\textbf{11.62}} \\

3
& 97.40 & 97.78
& 97.76 & 99.05
& 99.06 & 98.60
& 60.47 & 97.26
& 63.44 & 96.09
& 78.44 & 94.46
& 98.40 & 98.48
& 67.45 & \cellcolor{gray!20}{\textbf{95.94}}
& 30.95 & \cellcolor{gray!20}{\textbf{2.54}} \\

4
& 97.40 & 98.52
& 80.12 & 98.08
& 96.90 & 98.73
& 68.72 & 96.42
& 98.43 & 98.88
& 98.13 & 99.21
& 97.71 & 98.84
& 74.42 & \cellcolor{gray!20}{\textbf{97.25}}
& 23.29 & \cellcolor{gray!20}{\textbf{1.59}} \\

5
& 97.92 & 98.77
& 79.18 & 97.67
& 96.90 & 98.10
& 99.12 & 99.57
& 99.69 & 97.77
& 96.63 & 99.21
& 98.05 & 98.68
& 79.18 & \cellcolor{gray!20}{\textbf{97.67}}
& 18.87 & \cellcolor{gray!20}{\textbf{1.01}} \\

6
& 98.18 & 98.77
& 100.00 & 98.97
& 98.14 & 97.78
& 97.80 & 99.57
& 97.17 & 98.60
& 98.50 & 95.63
& 98.30 & 98.22
& -- & \cellcolor{gray!20}{--}
& -- & \cellcolor{gray!20}{--} \\
\hline
\end{tabular}
\end{table*}

\begin{table}[t]
\centering
\caption{Comparison between Ours and Baseline across different numbers of training positions.}
\label{tab:CrossPosition2}
\renewcommand{\arraystretch}{1.2}
\setlength{\tabcolsep}{4pt}

\begin{tabular}{c | c c c c c}
\hline
$|\mathcal{P}_{train}|$ & 1 & 2 & 3 & 4 & 5 \\
\hline
Ours Unseen (\%) & \cellcolor{gray!20}{\textbf{80.35}} & 87.20 & 95.94 & 97.25 & 97.67 \\
Baseline Unseen (\%) & 60.57 & 67.97 & 67.45 & 74.42 & \cellcolor{gray!20}{\textbf{79.18}} \\
\rowcolor{gray!20}
$\Delta$\% (O-B) & \textbf{+19.78} & \textbf{+19.23} & \textbf{+28.49} & \textbf{+22.83} & \textbf{+18.49} \\
Baseline Pos Needed & 5 & --- & --- & --- & --- \\
Data Reduction & 5$\times$ & --- & --- & --- & --- \\
\hline
\end{tabular}
\end{table}

\section{Experiments}

\subsection{Experimental Setup}

\begin{figure*}[!t]
    \centering
    \includegraphics[width=18cm]{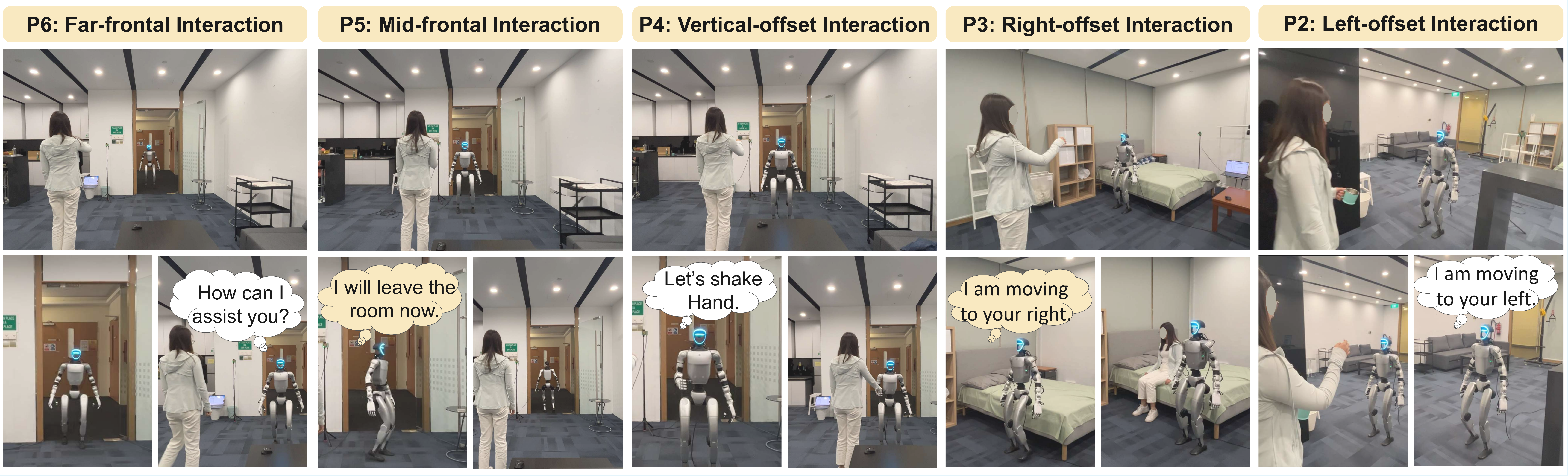}
    \caption{Illustration of room-scale humanoid–robot interaction scenarios under diverse user positions.}
    \label{fig_6_1}
\end{figure*}

\begin{figure}[!t]
    \centering
    \includegraphics[width=8.5cm]{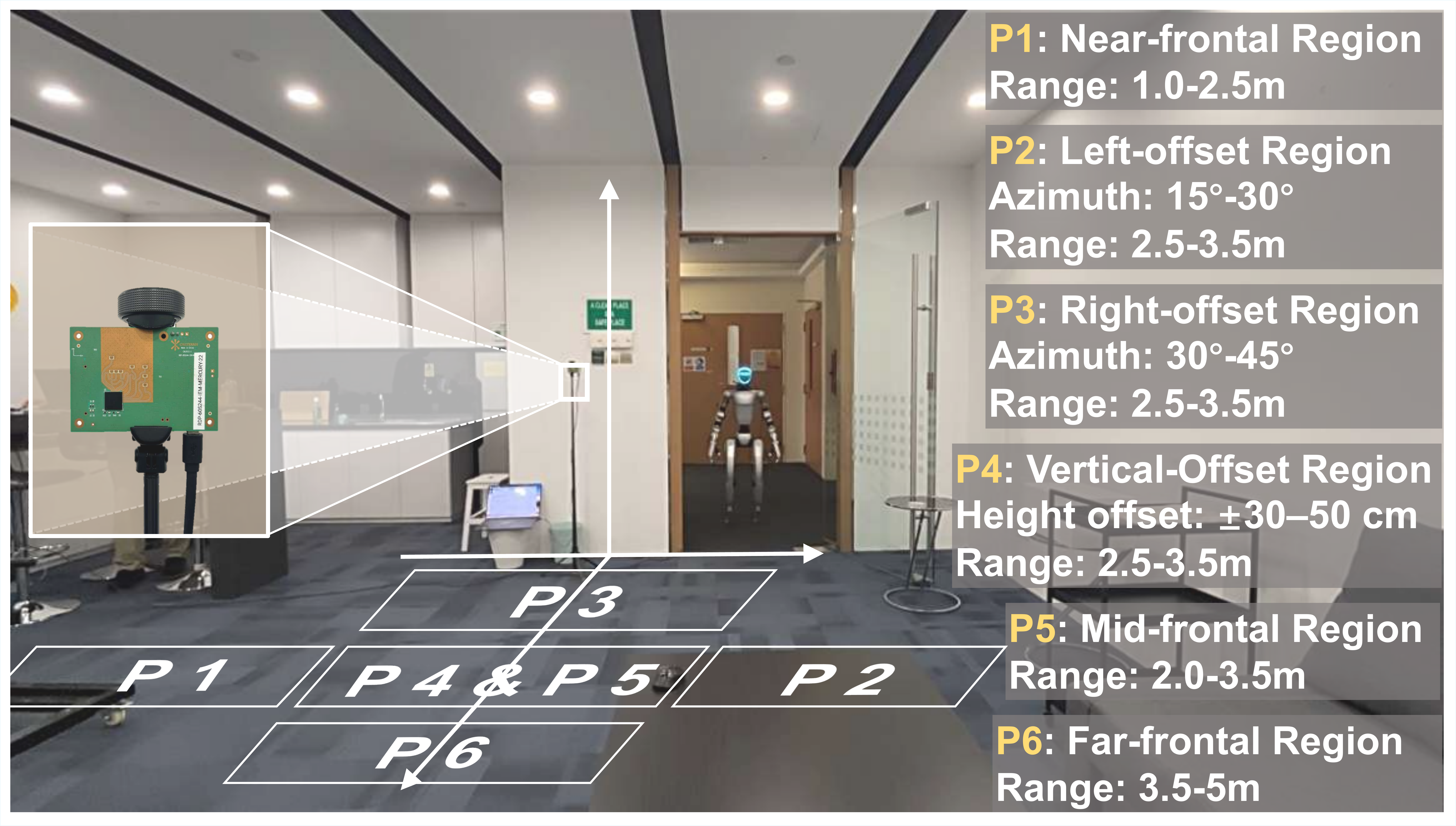}
    \caption{Illustration of the experimental environment and the six defined spatial positions ($\mathrm{P1}$-$\mathrm{P6}$) within the radar field of view.}
    \label{fig_6}
\end{figure}

\subsubsection{Hardware Platform}
All experiments are conducted on a robotic interaction platform that integrates a commercial FMCW mmWave radar and a Unitree G1 humanoid robot. 
The mmWave radar serves as the primary sensing module for room-scale human--robot interaction, while the G1 robot provides a human-like embodiment for executing interactive behaviors.

We employ a commercial FMCW MIMO radar (Calterah RDP60S244-IEM MERCURY) as the sensing front-end. 
The radar adopts a \(4 \times 4\) TX/RX MIMO configuration, forming 16 virtual channels for high-resolution angular perception, as illustrated in Fig.~\ref{fig_2}. 
It operates at a start frequency of 60.5~GHz with a 3.5~GHz bandwidth and a 10~MHz ADC sampling rate. 
Each frame contains 256 chirps, and each chirp includes 256 ADC samples. 
The FMCW slope is 118.24~MHz/\(\mu\)s, and the radar frame period is set to 50~ms. 
Under these settings, the radar achieves a maximum detection range of 6.4~m, a range resolution of 0.05~m, a maximum radial velocity of 6.4~m/s, and a velocity resolution of 0.05~m/s.

The perception outputs from the mmWave radar are transmitted to a Unitree G1 humanoid robot, which acts as the interactive agent throughout all experiments. 
During experiments, the robot receives real-time perception results for interpreting human interaction cues and executing responsive behaviors.

\subsubsection{Data Collection and Participants}
Gesture data are collected in an indoor room-scale environment containing six pre-defined spatial positions within the radar field of view, denoted as $\mathrm{P1}$–$\mathrm{P6}$ (Fig.~\ref{fig_6}). 
These positions correspond to different combinations of azimuth offsets, elevation offsets, and user-radar distances, enabling systematic evaluation of cross-position generalization.
Five participants perform five gesture classes at each position. 
Each gesture is repeated 80 times, producing a dataset of 12,000 samples 
(6 positions $\times$ 5 participants $\times$ 5 gestures $\times$ 80 repetitions).

\subsubsection{Training and Testing Protocol}
Unless otherwise specified, $\mathrm{P1}$ (or $\mathrm{P1}$–$\mathrm{P3}$ for multi-position training) is used as the seen training position and the remaining positions are treated as unseen during inference. 
All compared methods adopt identical training and testing splits to ensure fair comparison. 
Each model is trained for 100 epochs using only the training-position samples and evaluated on all six positions. 
Cross-position accuracy, unseen-position mean accuracy, and overall mean accuracy are reported in the following sections.

\subsubsection{Implementation Details}
Spectrogram enhancement and recognition networks are implemented in PyTorch and trained on a workstation equipped with an Intel Core i5-9500 CPU, 16~GB RAM, and an NVIDIA RTX~2070 GPU with 8~GB memory.
The complete inference pipeline, including enhancement, and classification, requires less than 6~ms per sample on this hardware configuration. 
All experiments are conducted under the same computational environment for consistent comparison.

\begin{figure}[!t]
    \centering
    \includegraphics[width=7.5cm]{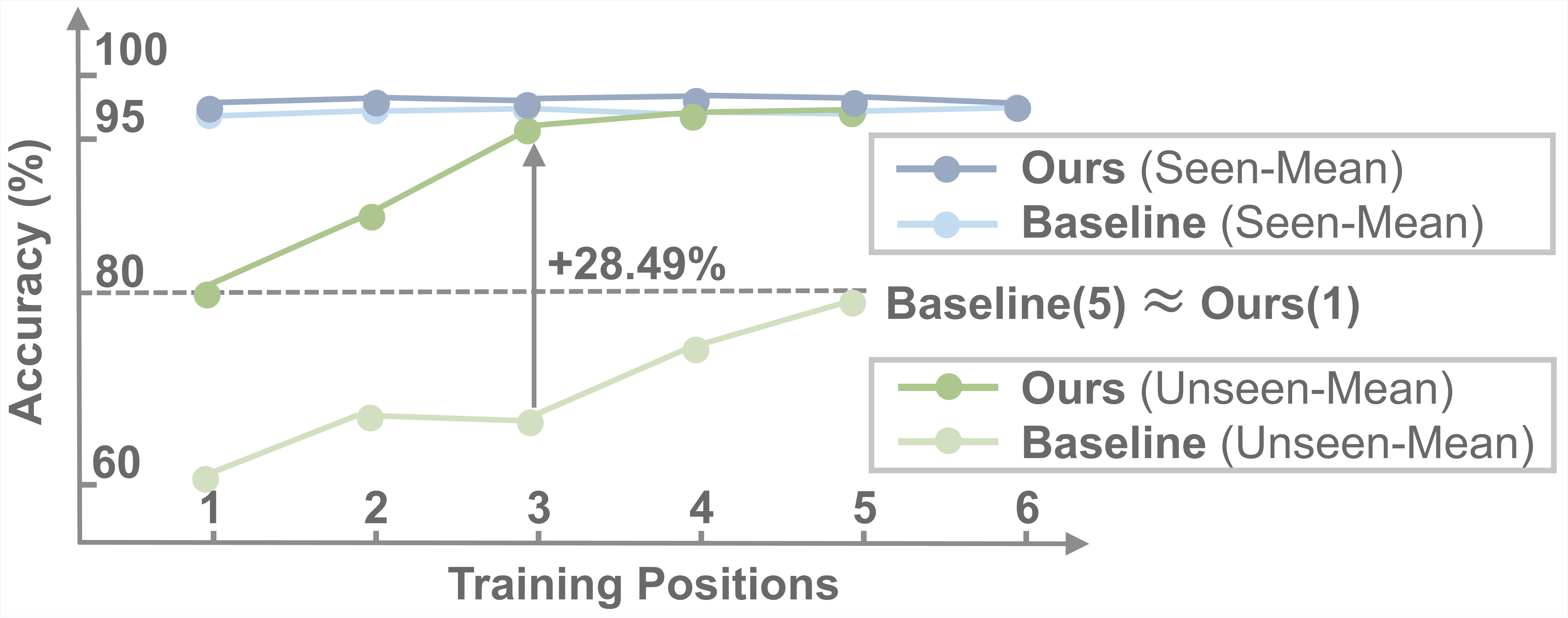}
    \caption{Seen-mean and unseen-mean accuracy trends for the baseline and the proposed method across different numbers of training positions.}
    \label{fig_7}
\end{figure}

\begin{figure}[!t]
    \centering
    \includegraphics[width=7.5cm]{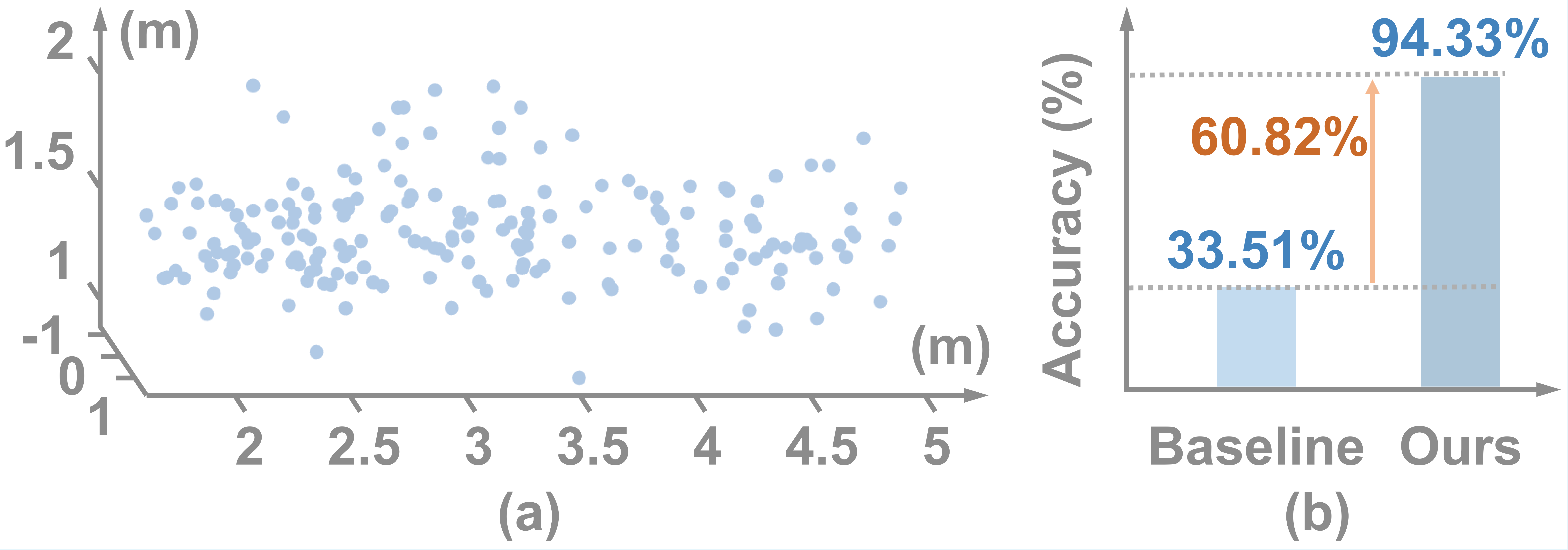}
    \caption{
    3D scatter distribution of each sample (a) and accuracy comparison under random-position testing (b). 
    }
    \label{fig_9}
\end{figure}

\begin{table*}[t]
\centering
\caption{
Ablation study under $|\mathcal{P}_{\mathrm{train}}|=3$ (training at $\mathrm{P1}, \mathrm{P3}, \mathrm{P5}$). 
Left: accuracy on training and unseen test positions. 
Right: unseen-position accuracy for $\mathrm{P2}, \mathrm{P4}, \mathrm{P6}$.
}
\label{tab:AblationStudy}
\makebox[\textwidth][c]{

\begin{minipage}{0.65\textwidth}
\renewcommand{\arraystretch}{1.20}
\setlength{\tabcolsep}{3pt}
\centering
\begin{tabular}{l|ccc|ccc|ccc}
\hline
\multirow{2}{*}{Method} 
& \multicolumn{3}{c|}{Training Pos.} 
& \multicolumn{3}{c|}{Test Pos.} 
& \multirow{2}{*}{Train-Mean $\uparrow$} 
& \multirow{2}{*}{Test-Mean $\uparrow$} 
& \multirow{2}{*}{Gap $\downarrow$} \\
& P1 & P3 & P5 & P2 & P4 & P6 & & & \\
\hline
Baseline (Raw Only) 
& 97.40 & 97.76 & 99.06 
& 60.47 & 63.44 & 78.44 
& 98.40 & 67.45 & 30.95 \\

w/ Alignment 
& 98.28 & 96.84 & 98.60 
& 97.12 & 95.75 & 89.87 
& 97.90 & 94.25 & 3.65 \\

w/ Alignment+Enh. 
& 97.29 & 98.42 & 97.77 
& 95.20 & 96.26 & 92.56 
& 97.82 & 94.67 & 3.15 \\

w/ Attention Only 
& 97.66 & 97.83 & 99.37 
& 72.12 & 68.54 & 75.75 
& 98.29 & 72.14 & 26.15 \\

\rowcolor{gray!20}
Full Ours (A+E+Attn) 
& 97.78 & 99.05 & 98.60 
& 97.26 & 96.09 & 94.46 
& 98.48 & 95.94 & 2.54 \\
\hline
\end{tabular}
\end{minipage}
\hfill
\begin{minipage}{0.32\textwidth}
\centering
\includegraphics[width=0.75\linewidth]
{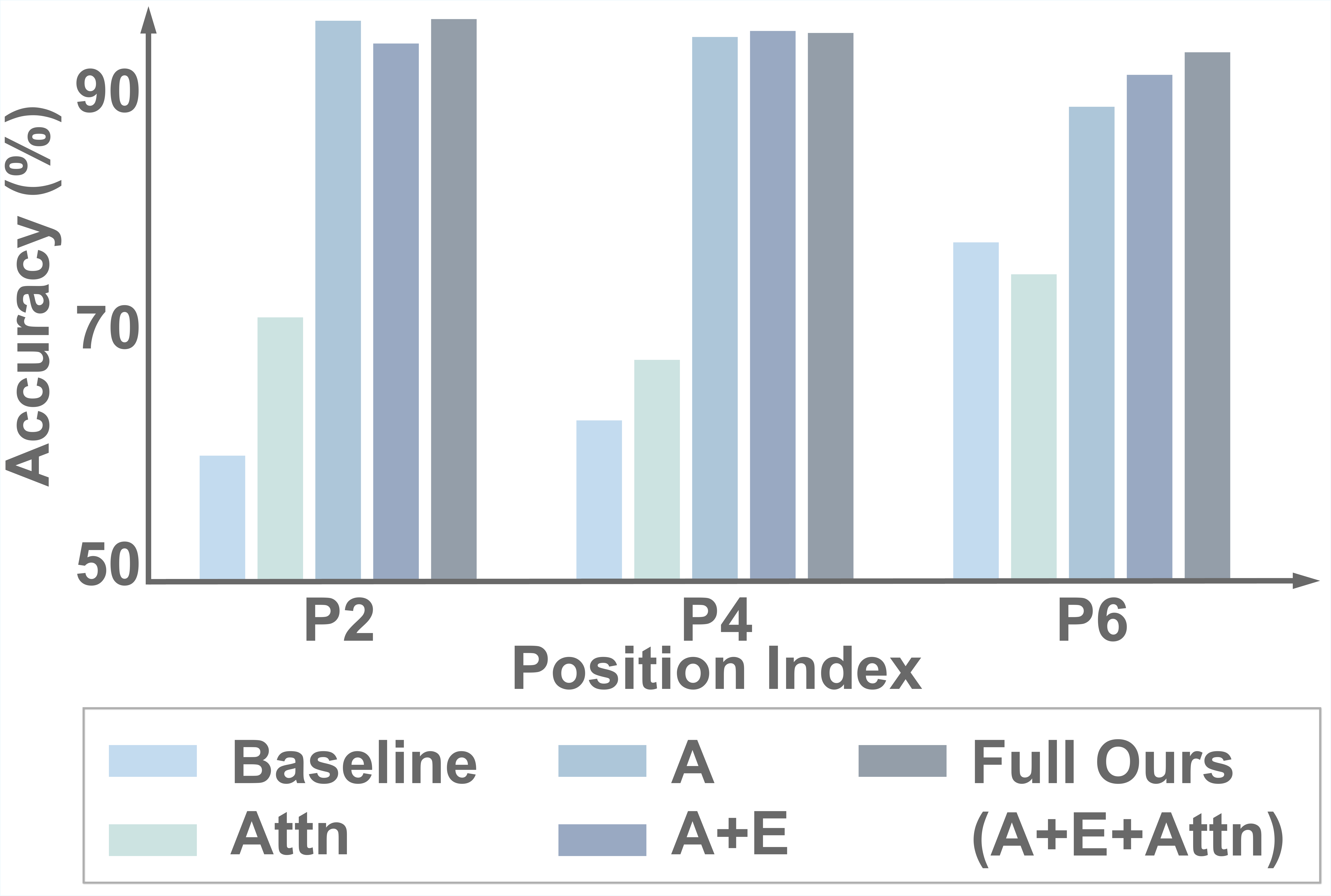}
\end{minipage}
}
\end{table*}

\subsection{Cross-Position Generalization Results}

\subsubsection{Fixed Unseen-Position Results}

Table~\ref{tab:CrossPosition1} and Table~\ref{tab:CrossPosition2} summarize the cross-position recognition accuracy when the model is trained on one to six spatial positions and evaluated on all six. Across all configurations, the proposed method significantly improves unseen-position accuracy compared with the baseline. When trained on a single position ($|\mathcal{P}_{train}|=1$), our method achieves 80.35\% unseen accuracy, outperforming the baseline by 19.78\%. As the number of training positions increases, unseen accuracy continues to improve and reaches 97.67\% with five training positions, while the baseline still remains below 80\%. This trend demonstrates that our spatial alignment and spectral enhancement modules effectively mitigate viewpoint- and distance-induced distortions and provide robust representations even under limited training diversity.

A position-by-position analysis reveals that the most challenging locations are typically those with large azimuth or elevation offsets (e.g., $\mathrm{P2}$, $\mathrm{P4}$) or long-range conditions (e.g., $\mathrm{P5}$, $\mathrm{P6}$). These positions cause severe geometric deformation or spectral sparsity for the baseline, with accuracy dropping to 50–70\%. In contrast, our method maintains consistently high accuracy across all unseen positions. For instance, at configuration $|\mathcal{P}_{\mathrm{train}}|=3$, accuracy at $\mathrm{P4}$ improves from 60.47\% to 97.26\%, and at $\mathrm{P5}$ from 63.44\% to 96.09\%. These results demonstrate that the proposed framework not only improves overall generalization but also stabilizes performance under challenging positional variations encountered in room-scale environments.

The overall trend is further illustrated in Fig.~\ref{fig_7}, which plots the mean seen and unseen accuracy as a function of the number of training positions. The baseline exhibits a clear dependence on training diversity: unseen accuracy increases slowly and only approaches 80\% when five training positions are provided. In contrast, the proposed method achieves strong generalization even under limited supervision, reaching above 80\% unseen accuracy with only a single training position and surpassing 95\% when three positions are used. The gain at $|\mathcal{P}_{train}|=3$ reaches 28.49\%, highlighting that geometric alignment and spectral enhancement effectively remove position-induced variations that otherwise require extensive multi-position training. This behavior indicates that our method learns position-invariant representations and drastically reduces the amount of data required for robust cross-position recognition.

\subsubsection{Random Free-Position Evaluation}

To further evaluate room-scale generalization beyond the predefined positions $\mathrm{P1}$-$\mathrm{P6}$, we conduct a random-position test in which users freely move within the radar coverage and perform gestures from arbitrary viewpoints. This scenario produces highly diverse combinations of distance, azimuth, and elevation, and therefore represents a more realistic and challenging setting. As shown in Fig.~\ref{fig_9}(b), the baseline achieves only 33.51\% accuracy due to severe geometric deformation and spectral sparsity. In contrast, the proposed method reaches 94.33\%, representing a substantial improvement of 60.82 percentage points. The 3D scatter comparison in Fig.~\ref{fig_9}(a) shows that the aligned point clouds form a compact and stable structure, enabling reliable recognition under unconstrained spatial configurations.

These results confirm that the proposed alignment and enhancement pipeline generalizes effectively to free-form, previously unseen positions and supports robust gesture recognition in room-scale scenarios where the user location cannot be predetermined. This ability is essential for humanoid-robot interaction, where users naturally move throughout the environment rather than standing at fixed viewpoints.

\begin{table*}[t]
\centering
\caption{
Cross-position gesture recognition accuracy (\%) for the proposed method and reproduced baseline approaches.
Training positions: $\mathrm{P1}$, $\mathrm{P3}$, $\mathrm{P5}$; testing positions: $\mathrm{P2}$, $\mathrm{P4}$, $\mathrm{P6}$.
}
\label{tab:cross_pos_full}
\renewcommand{\arraystretch}{1.20}
\setlength{\tabcolsep}{5pt}

\begin{tabular}{l c | ccc | ccc | cccc}
\hline
\multirow{2}{*}{Method} & \multirow{2}{*}{Input}
& \multicolumn{3}{c|}{Training Pos.} 
& \multicolumn{3}{c|}{Test Pos.}
& \multirow{2}{*}{Train-Mean $\uparrow$}
& \multirow{2}{*}{Test-Mean $\uparrow$}
& \multirow{2}{*}{Gap $\downarrow$}
& \multirow{2}{*}{Random-Mean $\uparrow$} \\
\cline{3-8}
 & & $\mathrm{P1}$ & $\mathrm{P3}$ & $\mathrm{P5}$ & $\mathrm{P2}$ & $\mathrm{P4}$ & $\mathrm{P6}$ \\
\hline

TS-DRSPA~\cite{20} & 3-channel spectrum
& 96.91 & 96.53 & 96.97
& 83.70 & 80.78 & 75.69
& 96.80 & 80.06 & 16.74 & 63.43 \\

DI-Gesture~\cite{310} & range-angle image
& 94.33 & 96.18 & 95.15
& 77.53 & 75.68 & 58.56
& 95.22 & 70.59 & 24.63 & 76.12 \\

ShuffleNet-Traj~\cite{341} & 3-channel spectrum
& 96.65 & 96.53 & 95.45
& 72.33 & 69.39 & 69.27
& 96.22 & 70.39 & 25.83 & 61.34 \\

5D-DCN~\cite{26} & 4-channel spectrum
& 94.85 & 94.44 & 93.33
& 86.99 & 71.94 & 65.60
& 94.21 & 74.84 & 19.37 & 73.72 \\

Ours-Basic & 3-channel spectrum
& 97.40 & 97.76 & 99.06
& 60.47 & 63.44 & 78.44
& 98.40 & 67.45 & 30.95 & 33.51 \\

\rowcolor{gray!20}
Ours & 3-channel spectrum
& 97.78 & 99.05 & 98.60
& 97.26 & 96.09 & 94.46
& 98.48 & 95.94 & 2.54 & 94.33 \\
\hline
\end{tabular}
\end{table*}

\subsection{Visual Results}
Fig.~\ref{fig_5} provides both qualitative and quantitative evidence that the proposed alignment and enhancement modules substantially improve the spatial coherence of mmWave spectrograms. These improvements extend beyond classification accuracy, as they directly strengthen the geometric and spectral consistency of radar measurements, which is critical for downstream spatially varying sensing tasks.

In the top row, raw spectrograms exhibit pronounced cross-position inconsistencies, where samples of the same gesture are dispersed across positions in the manifold visualization. After alignment, spectral shapes become more consistent and the manifold contracts into tighter clusters. All three quantitative metrics (IGD, CentroidSpread, and CSI) decrease accordingly, confirming that alignment better preserves the underlying spatial structure.
The bottom row illustrates the effect of spectral enhancement under long-range sparse conditions. Raw sparse spectrograms suffer from fragmented reflections and incomplete motion trajectories. After enhancement, energy continuity is restored and motion patterns become more discernible. The manifold visualization shows increased compactness, and the quantitative metrics further decrease, indicating reduced spectral inconsistency.

Specifically, in the cross-position setting, IGD decreases from 0.45 to 0.39, CentroidSpread from 0.38 to 0.19, and CSI from 0.23 to 0.14. In the long-range sparse case, IGD decreases from 0.47 to 0.37, CentroidSpread from 0.40 to 0.17, and CSI from 0.24 to 0.14, corresponding to reductions of approximately 20\%, 58\%, and 62\%, respectively. These results indicate that alignment primarily improves geometric compactness across viewpoints, while enhancement reinforces spectral continuity.
Together, these modules yield position-invariant and spectrally stable representations, suggesting their applicability beyond gesture recognition to broader spatial perception tasks that require viewpoint-robust sensing.

\subsection{Ablation Study}
To assess the contribution of each component, we conduct an ablation study with $|\mathcal{P}_{\mathrm{train}}|=3$ (training at $\mathrm{P1}$, $\mathrm{P3}$, and $\mathrm{P5}$). Results are reported in Table~\ref{tab:AblationStudy}, with the right-hand bar plot illustrating unseen-position accuracy at $\mathrm{P2}$, $\mathrm{P4}$, and $\mathrm{P6}$.
Alignment yields the largest single-component gain, improving unseen accuracy from 67.45\% to 94.25\% and reducing the generalization gap from 30.95\% to 3.65\%, confirming its critical role in correcting viewpoint-induced geometric distortion. Incorporating spectral enhancement further strengthens long-range spectral density, increasing the test-mean accuracy to 94.67\%. By contrast, attention alone provides marginal improvement, as geometric and spectral inconsistencies persist, leaving the gap (26.15\%) close to the baseline.
The full model integrating alignment, enhancement, and attention achieves the best performance, with \textit{Full Ours} reaching a test-mean accuracy of 95.94\% and a minimal gap of 2.54\%. The bar plot shows consistent gains across all unseen positions, particularly at $\mathrm{P2}$ and $\mathrm{P4}$ where raw spectrograms suffer from severe geometric distortion and energy sparsity. These results demonstrate that the three modules play complementary roles in enabling stable viewpoint-invariant gesture recognition.

\subsection{Comparison with Existing Methods}

To ensure a fair and representative comparison, we selected several recent gesture recognition methods that explicitly account for spatial variation and reproduced their architectures according to the original papers. All methods were trained and evaluated under the same cross-position protocol, with $\mathrm{P1}$, $\mathrm{P3}$, and $\mathrm{P5}$ used for training and $\mathrm{P2}$, $\mathrm{P4}$, and $\mathrm{P6}$ for unseen-position testing. This protocol isolates spatial generalization from dataset bias. Detailed per-position results and aggregated metrics (Train-Mean, Test-Mean, generalization gap, and random-position accuracy) are reported in Table~\ref{tab:cross_pos_full}.
Overall, our method demonstrates the strongest generalization among all approaches. While all baselines achieve comparable performance on training positions (Train-Mean $>$ 94\%), their accuracies degrade substantially on unseen positions. For instance, ShuffleNet-Traj and DI-Gesture reach Test-Mean values of only 70.39\% and 70.59\%, with large generalization gaps of 25.83 and 24.63 percentage points, respectively. Even the 5D-DCN model, despite leveraging four-channel spectrum inputs, still exhibits a notable gap of 19.37.
By contrast, our approach maintains consistently high performance across both training and unseen positions, achieving a Test-Mean of 95.94\% with an exceptionally small gap of 2.54. Moreover, the random-position mean accuracy further confirms the robustness of our model: whereas existing methods range from 61\% to 76\%, our framework attains 94.33\%, indicating strong stability under arbitrary spatial perturbations. These results verify that the proposed alignment, enhanced spectral representation, and attention modeling effectively improve spatial generalization in room-scale gesture recognition.

\section{Conclusion}

This work presented a spatially adaptive gesture recognition framework for room-scale human--humanoid interaction. By jointly addressing geometric distortion and spectral sparsity, the proposed alignment and enhancement modules produce spatially consistent representations that generalize across user positions. As a result, unseen-position accuracy improves from 67.45\% to 95.94\%, and random free-position accuracy increases from 33.00\% to 94.33\%, accompanied by substantial reductions in CentroidSpread and CSI of up to 58\% and 62\%. These results demonstrate robust cross-position sensing beyond fixed-view configurations and indicate applicability to broader mmWave-based perception tasks. Building on this capability, future work will explore multimodal humanoid robot perception by integrating mmWave sensing with complementary modalities, leveraging public datasets such as the mm-Fi dataset.


\bibliographystyle{IEEEtran}
\bibliography{Bibliography/BIB}\ 


\end{document}